\documentclass[10pt,twocolumn,letterpaper]{article}

\usepackage{iccv}
\usepackage{times}
\usepackage{epsfig}
\usepackage{graphicx}
\usepackage{amsmath}
\usepackage{amssymb}
\usepackage{comment}

\usepackage[pagebackref=true,breaklinks=true,letterpaper=true,colorlinks,bookmarks=false]{hyperref}

\iccvfinalcopy 

\def\iccvPaperID{42} 

\ificcvfinal\fi

\newcommand{\convnets}{convolutional neural networks\xspace}
\newcommand{\stn}{STNet\xspace}

\begin{document}

\title{Compact Neural Representation Using Attentive Network Pruning}

\author{Mahdi Biparva, John Tsotsos\\
	Department of Electrical Engineering and Computer Science\\
	York University\\
	Toronto, Canada\\
	{\tt\small \{mhdbprv,tsotsos\}@eecs.yorku.ca}}

\maketitle
\ificcvfinal\thispagestyle{empty}\fi

\begin{abstract}
	Deep neural networks have evolved to become power demanding and consequently difficult to apply to small-size mobile platforms. Network parameter reduction methods have been introduced to systematically deal with the computational and memory complexity of deep networks. 
We propose to examine the ability of attentive connection pruning to deal with redundancy
reduction in neural networks as a contribution to the reduction of computational demand. In this work, we describe a Top-Down
attention mechanism that is added to a Bottom-Up feedforward network to select
important connections and subsequently prune redundant ones at all parametric layers. Our method not only introduces a novel hierarchical selection mechanism as the basis of pruning but also remains competitive with previous
baseline methods in the experimental evaluation. We conduct experiments
using different network architectures on popular benchmark datasets
to show high compression ratio is achievable with negligible loss of accuracy.

\end{abstract}

\section{Introduction}
\label{sec:int}
The human brain receives a tremendously large amount of raw sensory
data at every second. How the brain deals efficiently and accurately with
the amount of input data to accomplish short- and long-range tasks
is the target of various research studies. \cite{tsotsos1990analyzing,tsotsos1995SelTun}
analyze the computational complexity of visual tasks and suggest the
brain employs approximation solutions to overcome some of the difficulties presented due to the vast amount of input sensory data. 

Neural networks have been successful on various computational tasks
in vision, language, and speech processing. Such networks are defined
using a large number of parameters arranged in multiple layers of
computation. Despite achieving good performance on
benchmark datasets, parametric redundancy is known to be widespread,
hence not suitable for real-time mobile applications. Tensor processing, memory usage, and power consumption of mobile devices are limited and consequently neural networks must be accelerated and compressed for such mobile applications \cite{han2015deep}. Model compression reduces the number of parameters and primitive operations and consequently improve the computation speed at inference phase \cite{han2015deep}.

Moreover, due to the over-fitting phenomenon, over-parameterized models suffer from low generalization and therefore must be regularized. Such models learn dataset biases very quickly, memorize data distribution, and consequently lack proper generalization. One way to regularize parametric models is by imposing sparsity-imposing terms and consequently pruning a number of parameters to zero and keeping a sparse subset of them \cite{wen2016struspar,zhou2016less}.

Neural network compression is defined as a systematic attempt to reduce parametric redundancies in dense and multi-layer networks while maintaining generalization performance with the least accuracy drop. Parametric redundancy in such networks is empirically investigated in \cite{denil2013predicting}.
Various neural network compression approaches such as 
weight clustering \cite{han2016deep,chen2015compressing,park2017weighted}, 
low-rank approximation \cite{denil2013predicting,denton2014exploiting}, 
weight pruning \cite{han2015learning,guo2016dynamic,dong2017learning,han2016deep,luo2017thinet,lecun1990optimal,hassibi1993second},
and sparsity via regularization \cite{liu2015sparse,zhou2016less,wen2016learning}
are introduced to reduce parameter redundancy for lower computational and memory complexity. 
Pruning methods have shown to be computationally favorable while relying on straightforward heuristics and ad hoc approaches to schedule and devise pruning patterns. These compression approaches rely on defining some measure of importance based on which a significant subset of weight parameters are kept and the rest are pruned permanently.



STNet \cite{biparva2017stnet} introduces a selective attention approach in \convnets for the task of object localization. STNet leverages a small portion of the entire visual hierarchy to route through all parametric layers to localize the most important regions of the input images for a top label category. The selection process is hierarchical and provides a reliable source of weight pruning. The experimental results of STNet for object localization reveal that a sparse subset of the network units and weight parameters are sufficient for a successful localization result.
We propose a novel attentive pruning method based on STNet to achieve compact neural representation 
using Top-Down selection mechanisms. Following \cite{biparva2017stnet}, we define
a neural network to benefit from two passes of information processing,
Bottom-Up (BU) and Top-Down (TD). The BU pass is data-driven. It begins
from raw input data, goes through multiple layers of feature transformation,
and finally predicts abstract task-dependent outputs. On the other
hand, the TD pass is initialized from high level attention signals, goes
through selection layers, and outputs kernel importance activities.
The importance activities are computed by three variable inputs in
the TD selection pass: one is the hidden responses, the other is the
kernel filters, the last is the top attention signals. We show that
all of the three sources of TD selection are crucial for strong network
pruning.

Attentive pruning relies on kernel importance activities to decide
on pruning patterns. We feed neural networks with input data, and
then activate TD selection to output kernel importance activities
at every layer. These activities are accumulated and scheduled to
generate pruning patterns. We evaluate the attentive pruning method
using various network architectures on benchmark datasets. The competitive
evaluation results reveal the selective nature of the TD mechanisms
over the kernel filters. This complements the importance of the gating
activities for visual tasks such as object localization and segmentation.


\section{Network Pruning Using Selective Attention}
\label{sec:mod}

We define neural networks with Bottom-Up (BU) and Top-Down (TD) information passes. The former transforms input data into high-level semantic information. On the other hand, the TD pass begins from class predictions and computes the kernel importance responses at each layer. The TD selection process relies on three main sources of information. We propose to compute the important connections along which the TD attentional traces route through the visual hierarchy. In this work, we introduce a novel approach that the pruning mechanism relies on the accumulated kernel importance responses while the baseline models solely relies on kernel filters of the feedforward pass. The kernel importance responses are computed using the local competitions that receives three variable inputs: the kernel weights, the hidden activities, and the gating activities. The kernel weights are learned in the pre-training phase. The hidden activities represent the hierarchical feature representation of the underlying layers for some specific input data. Therefore, the calculated kernel importance responses take into consideration not only the kernel weights but also the input hidden activities. The baseline pruning methods only rely on the magnitude thresholding while the proposed method generalizes the baselines with the inclusion of the hidden activities. Furthermore, the kernel importance responses are category-specific. The important subset of weight parameters are determined not only for some specific input but also some particular label category. The TD selection pass starts from a category initialization signal. Consequently, all the TD selection mechanisms are informative of that particular category initialization. The category-specific nature of the attentive pruning method further reduce the non-relevant pruning and therefore, speeds up the convergence in the retraining phase. 

\begin{figure*}[t]
	\centering
	\includegraphics[width=2.1\columnwidth]{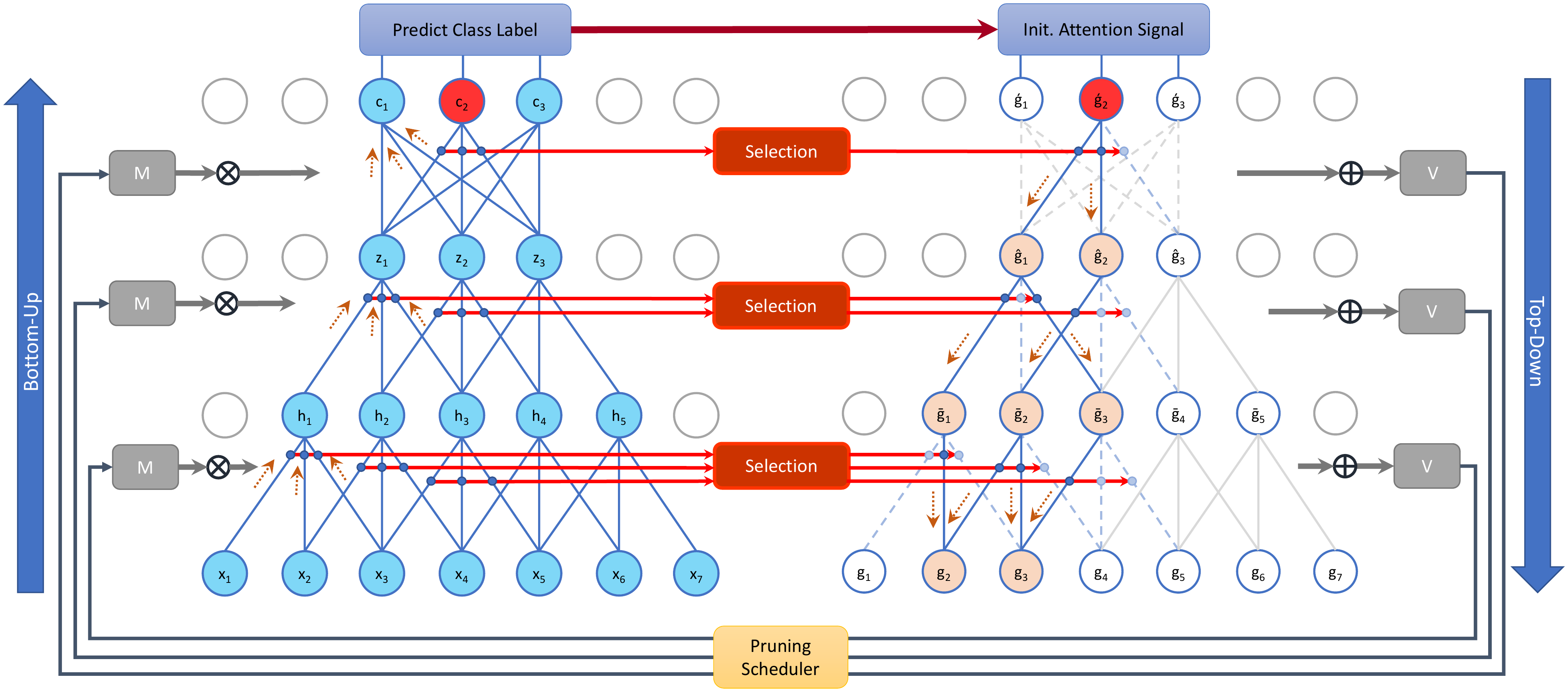}
	\caption{Schematic illustration of the proposed method for connection pruning that leads to the reduction of the number of network parameters. On the left side, a toy multi-layer feedforward network is shown. On the right, the corresponding TD networks is given. At each layer, once the active connections $\tilde{w}$ are computed using the TD selection mechanisms, they are additively accumulated into the persistent buffer $V$; subsequently, the mask tensor M is scheduled to get updated after a number of iterations. The feedforward pass is always additively modulated with the mask tensors M. The arrows show the direction of information flow.\label{fig:snp-overall-arch}}
\end{figure*}

\subsection{Method Overview}
Figure \ref{fig:snp-overall-arch} demonstrates schematically the information flow at different computational stages of the proposed method. First, given some input $x$ at the bottom of the visual hierarchy shown on the left part of Fig. \ref{fig:snp-overall-arch}, the feature extraction is done using the parametric layers and the output hidden activities $h$, $z$, $c$ are computed at each layer until the top score layer is reached and the BU pass ends. The Predict Class Label block is a multi-class transfer function such that $softmax()$ outputs the class probability prediction given the input data. Then, the attention signal initialization determines the label category for which the TD pass (shown on the right side) will be activated. Once the attention signal is set, the selection mechanism within the local receptive field of the initialized category node is activated. According to the competition result, a number of important outgoing connections on the TD layer are activated. Then, the gating activity of the category node proportional to the activated connection weights propagates downward to the next gating layer. This is illustrated by the outgoing solid (activated) and dashed (deactivated) connections from $\acute{g}$ to $\hat{g}$ in Fig. \ref{fig:snp-overall-arch}.
At this stage, the kernel importance responses $V$ for the top layer are updated with the activated connection patterns in an additive manner. This layer-wise computation continues at all of the subsequent lower layers until the TD selection pass ends at the input layer and returns the gating activities $g$. The kernel importance accumulation is iterated for a number of randomly-chosen input samples until a pruning phase is set by the scheduling strategy depicted by the yellow Pruning Scheduler module at the bottom of the figure. The pruning mask $M$, then, is updated based on the so-far-accumulated kernel importance responses. The pruning masks are initially set to one, meaning no kernel weight is pruned before the first scheduled mask update. Over different pruning phases, they start to gradually become zero.

\subsection{Notations}

A multi-layer neural network $f:\mathbb{R}^{H\times W}\rightarrow\mathbb{R}$
is at the core of the BU pass. The training set $D=\{(x_{i},y_{i})\}_{i=1}^{N}$ has $N$ samples such that the input data is $x\in\mathbb{R}^{H\times W}$an
input image with height $H$ and width $W$ and $y\in\{0,1,\dots,K-1\}$
is the ground truth label for $K$ different classes. We define the
BU pass as a feedforward network

\begin{equation}
	h=f(x;W),
\end{equation}
in which $f=\{f_{j}\}_{j=0}^{L}$ is a network with $L$ layers, $x$
and $h$ are the input and output of the network, and $W=\{W_{j}\}_{j=0}^{L}$
is the set of network parameters at $L$ layers. We define the feature
transformation $h_{l}=f(h_{l-1};W_{l})$ at layer $l$ such that $f(x;W)=W^{T}x$
is a linear transformation $f:\mathbb{R}^{M}\rightarrow\mathbb{R}^{N}$
of the input $x$ by the weight matrix $W\in\mathbb{R}^{M\times N}$
for fully-connected layers. The convolutional layers apply convolutions
using the kernel filter $W$.

The BU network output $h$ is fed into a Softmax transfer function
$\hat{y}=softmax(h)$ to compute the multinomial probability values
$\hat{y}$. The cross-entropy loss function is used with the Stochastic Gradient Descent (SGD) optimization algorithm to update network parameters.

\subsection{Top-Down Processing}

The role of the Top-Down (TD) pass is to traverse downward into the
visual hierarchy by routing through the most significant weight connections
of the network. TD pass begins from an initialization signal $d\in\mathbb{R}^{K}$
generated according to the ground truth label $y$. It traverses down layer
by layer by selecting through network connections

\begin{equation}
g=t(d,h,W),
\end{equation}
in which $h=\{h_{i}\}_{i=0}^{L}$ is the set of BU hidden activities,
network parameters $W=\{W_{i}\}_{i=0}^{L}$, and $g=\{g_{i}\}_{i=0}^{L}$
is the set of kernel importance responses at $L$ layers. The attention signal is initialized based on the ground truth label of the input image. It sets the
signal unit for the category label corresponding to the ground truth to one and keep the rest zero

\[
d=\{d_{j=y}=1,d_{j \neq y}=0,\}.
\]
The TD pass at each layer computes the importance responses using
three computational stages defined in STNet
\cite{biparva2017stnet}. STNet originally is used to compute attention maps for the gating activities while in this work, we seek to compute attention maps for the kernel parameters. 

The selection mechanism at each layer has three stages of computation. Each stage uses the element-wise multiplication of the hidden activities inside the receptive field of each top unit with the kernel parameters and then at the end of the selection mechanism propagates top gating activities proportional to the selected activities to the layer below. We review the three stages in the following based on the definitions in STNet. In the first stage, noisy redundant activities that interfere with the important activities are pruned away. The goal is to find the subset of the most critical activities that play the important role in the final decision of the network. 
In the second stage, among the selected activities returned by the first stage, activities are partitioned into connected components. This helps to impose the connectivity criterion that is crucial for a reliable hierarchical representation. The most informative group of activities are marked as the winner of the selection process at the end of the second stage. The local competition between groups is based on a combination of the size and total strength objectives. 
In the third stage, the selected winner activities are normalized such that they sum to one. Then, the the top gating activity is propagated proportional to the selected activities to the bottom gating layer, the activities of the bottom gating nodes are updated consequently. This procedure is repeated for a number of layer starting from the top of the network down to the early layers.

\subsection{Kernel Importance Maps \label{subsec:kim}}

The output hidden activities at a layer are computed by a linear multiplication
of the kernel weight matrix and the input hidden activities $h=W^{T}x$
for fully-connected layers. The extension to convolutional layers
is straightforward using convolution operations and are ignored for the sake of brevity. Each output
unit $h_{i}$ receives a weighted sum of the input vector $x$ according
to the weight parameter vector $w_{i}$ .

The TD selection mechanism $a^{l-1}=t(g_{j}^{l},x^{l},w_{j}^{l})$
for the output unit $j$ operates on the input hidden activities $x^{l}$,
the weight parameters $w_{j}^{l}$ connecting all the input units
to unit $j$, and the input gating unit $g_{j}^{l}$. The selection mechanism $t$ is only executed
for non-zero $g^{l}$ units. The output of the selection mechanism
contains two entities $a^{l-1}=\{g^{l-1},\tilde{w}^{l}\}$, where $g^{l-1}$ is the output
gating activities which is the source of TD selection at the layer
below, and $\tilde{w}^{l}$ is the kernel importance responses at
layer $l$. Hereafter, we drop $l$ for sake of notation brevity.
The kernel importance responses are accumulated for all $N$ samples
in the training set and used in selective pruning for network compression.

We categorize all the previous pruning approaches as class-agnostic pruning methods
since they determine the connections to prune regardless of the target
categories they are interested in. Our proposed attentive pruning method, however, is class-specific since the TD pass begins from some class hypothesis
signal and routes through the network hierarchy accordingly. Therefore,
the computed kernel responses are representative of the subset of
the network connections that are most important for the true category
label predictions.

Additionally, network parameters are trained according to the input
data distribution. The BU information flows into the network hierarchy
by measuring the numeric relation between an input unit $x_{i}$
and a connection weight $w_{ji}$ that connects the input unit $i$
to the output unit $j$. If both the input and the weight have high
activity, the output will have high value too. Being motivated by
this insight, we show that the TD selection process produces kernel
importance maps by considering both of the input and the kernel weights.
Kernel importance is measured based on whether the input units and
the kernel weights are both positively related. This generalizes the
previous works in which the kernel weights are individually considered
for connection pruning.

\subsection{Attentive Pruning}

We define an attentive pruning method using the kernel importance responses
$\tilde{W}$. The importance responses $\tilde{W}^{t}$ at iteration
$t$ is accumulated into a persistent buffer $V^{t}=V^{t-1}+\tilde{W}^{t-1}$.
The binary pruning mask $m$ defines the pattern using which the kernel weights
are permanently pruned. The function $r$ determines the pruning mask
$m$. $r$ is a thresholding function that sets the binary values
of $m$: 

\begin{equation}
	r(u;a)=\begin{cases}
	0 & u\leq a\\
	1 & a<u
	\end{cases},
\end{equation}
where $a=m(u)+\lambda\,\sigma(u)$. $\lambda$ is a multiplicative
factor, $m(u)$ is the mean, and $\sigma(u)$ is the standard deviation
of the input $u$. We set the binary mask $m_{l}$ at layer $l$ by
setting $u=V_{l}^{t}$ and $m_{l}=r(u;a)$. We run the BU and TD passes
for a number of iterations after which the attentive pruning starts
to compress the network parameters. Once the set of mask binary tensors
$m=\{m_{i}\}_{i=0}^{l}$ are determined after each pruning phase,
the feedforward BU pass is modified using the binary pattern in the
mask tensors:

\begin{equation}
h=m\odot W^{T}x,
\end{equation}
where $a\odot b$ is the element-wise (Hadamard) product of $a$ with $b$.

\begin{figure*}[t]
	\centering
	\includegraphics[width=1.8\columnwidth]{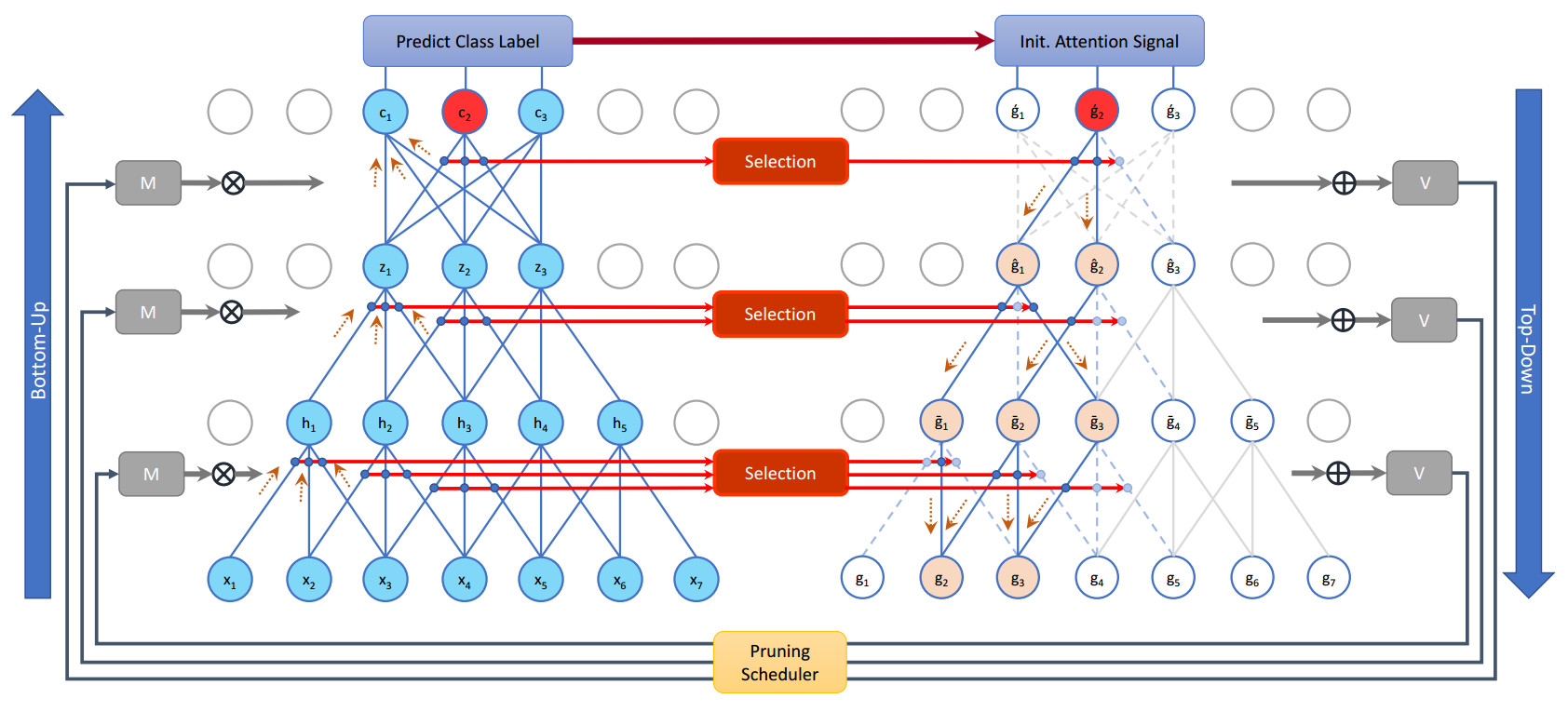}
	\caption{Detailed demonstration of different stages of computation of the BU and TD passes for selective connection pruning. At each layer, the inputs to the TD selection unit, the active connections $\tilde{w}$, the additive accumulation into the persistent buffer, and the multiplicative mask of the BU kernel weight are depicted.\label{fig:snp-modular}}
\end{figure*}

Fig. \ref{fig:snp-modular} illustrates the BU and TD interactions in detail. At the layer $i$ for instance, the TD selection mechanism receives the three inputs: the hidden activities $h_{i-1}$, the kernel weights $w_i$, and the gating activities $g_i$. Once the selection is completed, the kernel importance maps $\tilde{w}$ are set for the downward gating activity propagation. Additionally, they are used to additively update the persistent buffer $V_i$. The pruning pattern of the kernel weights $w_i$ is determined according to the binary pruning mask $M_i$. The mask is updated according to the pruning scheduler unit. Once the scheduler set the updating on, the thresholding function $r$ updates the mask binary elements given the input persistent tensor. This procedure is applied to every layer the pruning is defined to be applied.

\subsection{Retraining Strategy}

At every iteration, using the samples in the mini-batch, we have sequentially the following computational stages: a
feedforward BU pass, attention signal initialization, a TD pass, and
an updating of the persistent buffer $V$ using the kernel importance
responses $\tilde{W}$. After a number of initial iterations to accumulate kernel importance responses into the persistent buffer, we start
pruning the network connections for several times. The network is
retrained from the first occurrence of pruning onward. This helps
the network retain its level of accuracy for label prediction over
multiple stage of connection pruning. Retraining is inevitable due
to the high pruning rate of the network weight parameters. The network needs some iterations to shift its representational capability for a high level
of label prediction accuracy. The retraining allows the adaptation to the reduced parameter space. It follows the exact optimization settings used for the pre-training of the network prior to the network compression.

\section{Experimental Results}
\label{sec:exp}
In this section, we conduct experiments to evaluate the compression
ratio of the attentive pruning method. The compression ratio is defined as the ratio of the total number of mask units over the total number of the non-zero mask units (active connections). We use the Pytorch deep learning
framework \footnote{\href{https://pytorch.org/}{https://pytorch.org/}}
\cite{paszke2017automatic} to implement the model for the experiments
of this work. The layers of the TD pass are implemented using the
code provided by \footnote{\href{https://github.com/mbiparva/stnet-object-localization}{https://github.com/mbiparva/stnet-object-localization}}
\stn \cite{biparva2017stnet}. We choose the learning rate $10^{-3}$,
momentum $0.9$, weight decay $0.0005$, and mini-batch size $64$
for the SGD optimizer unless otherwise mentioned.

\begin{table}[h]
	\resizebox{0.95\columnwidth}{!}{
		\begin{tabular}{l|c|c|c}
			\hline 
			Model & Top-1 error & Parameters & Compression\tabularnewline
			\hline 
			LeNet-300-100-reference & 3.3\% & 267K & -\tabularnewline
			LeNet-300-100-pruned & 3.8\% & 5.2K & 58$\times$\tabularnewline
			\hline 
			LeNet-5-reference & 2.1\% & 83K & -\tabularnewline
			LeNet-5-pruned & 3.2\% & 4.7K & 102$\times$\tabularnewline
			\hline 
		\end{tabular}
	}
	\caption{LeNet error rate and compression ratio on MNIST dataset using the
		attentive connection pruning.\label{tab:LeNet-MNIST}}
\end{table}

\begin{table}
	\resizebox{0.95\columnwidth}{!}{
		
		\begin{tabular}{l|c|c|c}
			\hline 
			Model & Top-1 error & Parameters & Compression\tabularnewline
			\hline 
			Lenet-5-reference & 38.2\% & 83K & -\tabularnewline
			Lenet-5-pruned & 39.4\% & 8.3K & 10$\times$\tabularnewline
			\hline 
			CifarNet-reference & 30.4\% & 84K & -\tabularnewline
			CifarNet-pruned & 31.1\% & 7.6K & 11$\times$\tabularnewline
			\hline 
			AlexNet-reference & 23.5\% & 390K & -\tabularnewline
			AlexNet-pruned & 24.8\% & 13K & 29$\times$\tabularnewline
			\hline 
		\end{tabular}
		
	}
	
	\caption{LeNet and CifarNet error rate and compression ratio on CIFAR-10 dataset
		using the attentive connection pruning.\label{tab:LeNet-CIFAR}}
\end{table}

\begin{table*}[t]
	\centering
	\resizebox{1.8\columnwidth}{!}{
		\begin{tabular}{lcccc}
			\hline 
			Method & Network & Dataset & Error-degradation & Compression Ratio\tabularnewline
			\hline 
			Han et al. \cite{han2015learning} & LeNet-300-100 & MNIST & 0.19\%  & 12$\times$\tabularnewline
			Guo et al. \cite{guo2016dynamic} & LeNet-300-100 & MNIST & 0.23\% &  56$\times$\tabularnewline
			Dong et al. \cite{dong2017learning} & LeNet-300-100 & MNIST & 0.20\% & 66$\times$\tabularnewline		    
			Ours &  LeNet-300-100 & MNIST & 0.50\% & 58$\times$\tabularnewline		    
			\hline 
			Han et al. \cite{han2015learning} & LeNet-5 & MNIST & 0.09\% & 12$\times$\tabularnewline
			Guo et al. \cite{guo2016dynamic} & LeNet-5 & MNIST & 0.09\% & 108$\times$\tabularnewline
			Dong et al. \cite{dong2017learning} & LeNet-5 & MNIST & 0.39\% & 111$\times$\tabularnewline				
			Ours &  LeNet-5 & MNIST & 0.90\% & 102$\times$\tabularnewline		    
			\hline 
		\end{tabular}
	}
	\caption{Comparison of the Compression ratio of the proposed method with the baseline approaches using LeNet-300-100 and LeNet-5 network architectures on MNIST. Error degradation is the difference between the original error and the error at the end of the retraining phase.\label{tab:compression-sota}}
\end{table*}

\subsection{The MNIST Dataset}

One of the popular datasets widely used in the machine learning community
to experimentally evaluate novel methods is MNIST dataset. It contains
gray-scale images of handwritten digits and is used for category classification.

We define the BU network for the MNIST dataset according to two classic network architectures:
LeNet-300-100 \cite{Lecun1998} and LeNet-5 \cite{Lecun1998}. The
former consists of three fully-connected layers with output channel
sizes of 300, 100, 10 successively and contains 267K learnable parameters.
We train it for 10 epochs to obtain the reference model for the BU
network. Lenet-5, on the other hand, has two convolutional layers
at the beginning. Similarly, it is trained for 10 epochs. It has 431K
learnable parameters.

After the first epoch that the persistent buffers are accumulated, we start pruning the network connections for the next 7 consecutive epochs.
The multiplicative factor $\lambda$ is set to the following values $[0.0,0.5,1.0,1.5,2.0,2.5,3.0]$.
We continue retraining the network for 25 epochs after the final
pruning stage. We follow this pruning protocol for both of the LeNet architectures.
Error rate and the compression ratio of the networks are given in
Table \ref{tab:LeNet-MNIST}. The results confirm the selectivity
nature of the TD mechanism in the parameter space of the BU network.
According to the experiment results, the kernel importance responses are shown a reliable source of connection
pruning using LeNet architectures on MNIST. The proposed model is capable of reducing the number of kernel weights 58 and 102 times for LeNet300 and LeNet-5 respectively for negligible performance accuracy drops.

We compare the compression performance of the proposed attentive pruning mechanism with the baseline approaches on MNIST in Table \ref{tab:compression-sota}. The experimental results reveal that the proposed approach outperforms two of the baseline approaches \cite{han2015learning,guo2016dynamic} using the LeNet architectures while remain competitive with \cite{dong2017learning}. It should be noted that \cite{dong2017learning} uses the computationally expensive second order derivatives of a layer-wise error function to derive the pruning policy while we only rely on the important kernel responses derived from the TD selection mechanisms. \cite{dong2017learning} exhaustively relies on second-order derivatives at each layer while we chose to determine kernel responses in a hierarchical manner. However, the proposed method can outperform \cite{han2015learning,guo2016dynamic} that use magnitude pruning of weight parameters. This supports the role of the TD selection mechanisms to determine the most important parameters of neural networks as the source of a pruning procedure.

\subsection{The CIFAR Dataset}

CIFAR-10 dataset \cite{krizhevsky2009learning} contains RGB images of the same size and scale as MNIST dataset. The dataset consists of natural images of 10 semantic categories for object classification. In comparison with MNIST, the goal is to benchmark classifier performance on a higher level of complexity using CIFAR-10. We evaluate the performance
of the proposed method using three network architectures on this dataset:
LeNet-5, CifarNet and AlexNet. CifarNet\footnote{\href{https://code.google.com/archive/p/cuda-convnet/}{https://code.google.com/archive/p/cuda-convnet/}}
\cite{krizhevsky2009learning} is a multi-layer network with three
convolutional layer and two fully-connected layers. It has larger
number of parameters than LeNet-5. AlexNet \cite{krizhevsky2012imagenet}
has 5 convolutional and 2 fully-connected layers.

We empirically choose a slightly different pruning and re-training policy for
the CIFAR-10 dataset since it has a lot more complexity and care must
be taken for connection pruning. First, we change the mini-batch size
to 16. The multiplicative factor $\lambda$ is set only to $0.5$.
However, unlike the MNIST pruning protocol, we prune layers individually. 
We observed in the preliminary experiments that this approach helps maintain the label prediction accuracy with the minimal performance compromise while keep the compression ratio high.
This policy helps the network to maintain its representation capability for
the classification task and avoid deteriorating learning collapses. We first
accumulate the kernel importance responses in the persistent buffer
for one epoch. Next, for every 4 epochs, we prune the connections of one
layer starting from the first parametric layer at the bottom to the last one at
the top of the BU network. Once the pruning of the last layer is done,
we continue re-training of the pruned network for 40 epochs and then
report the compression ratio in Table \ref{tab:LeNet-CIFAR}. For
all of the three networks, the attentive pruning method is able to
maintain the reference network error rate and achieve high compression
ratio.

\section{Conclusion}
\label{sec:con}
We propose a novel pruning method to reduce the number of parameters of multi-layer neural networks. The attentive pruning method relies not only a feedforward
feature representation pass but also a selective top-down pass. The
TD pass computes the most important parameters of the kernel filters
according to a selected category label. Additionally, the hidden activities
at each layer participate in the stages of the TD selection mechanism.
This ensures both the top semantic information and input data representation
play roles in the stages of kernel importance computation. We evaluate
the compression ratio of the proposed method on two classification
datasets and show the improvement on three popular network architectures. The network achieves a high compression ratio with minimal compromise of generalization performance.

{\small
	\bibliographystyle{ieee}
	\bibliography{thesis_phd}
}

\end{document}